\documentclass[11pt]{article}

\usepackage{booktabs}
\usepackage{multicol, multirow}
\usepackage{stfloats}
\usepackage{subcaption}

\usepackage{acl} %For final version

\usepackage{times}
\usepackage{amsmath, amssymb, amsthm}
\usepackage{latexsym}

\usepackage[T1]{fontenc}
\usepackage[utf8]{inputenc}
\usepackage[normalem]{ulem}

\usepackage{microtype}

\usepackage{inconsolata}

\usepackage{graphicx}

\usepackage[most]{tcolorbox}
\usepackage{listings}
\usepackage{xcolor}
\usepackage{enumitem}

\newcommand{\cZ}{\mathcal{Z}}
\newcommand{\topoframework}{\texttt{HalluZig}}

\newcommand{\cancel}[1]

\title{\texttt{HalluZig}: Hallucination Detection using Zigzag Persistence}

\author{Shreyas N. Samaga \hspace{0.2in} Gilberto Gonzalez Arroyo \hspace{0.2in} Tamal K. Dey \\
        Department of Computer Science \\ 
        Purdue University 
        \\ West Lafayette, IN \\
        \texttt{\{ssamaga,gonza982,tamaldey\}@purdue.edu}}

\begin{document}
\maketitle
\begin{abstract}
The factual reliability of Large Language Models (LLMs) remains a critical barrier to their adoption in high-stakes domains due to their propensity to hallucinate. Current detection methods often rely on surface-level signals from the model's output, overlooking the failures that occur within the model's internal reasoning process. In this paper, we introduce a new paradigm for hallucination detection by analyzing the dynamic topology of the \emph{evolution of model's layer-wise attention}. We model the sequence of attention matrices as a \emph{zigzag graph filtration} and use \emph{zigzag persistence},  a tool from Topological Data Analysis, to extract a topological signature. Our core hypothesis is that factual and hallucinated generations exhibit distinct topological signatures. We validate our framework, \topoframework{}, on multiple benchmarks, demonstrating that it outperforms strong baselines. Furthermore, our analysis reveals that these topological signatures are generalizable across different models and hallucination detection is possible only using structural signatures from partial network depth.
\end{abstract}

\section{Introduction}
\label{sec:intro}
Large Language Models (LLMs) form the foundation of modern natural language processing, powering systems for search, question answering, decision support for various domains such as healthcare, law and finance. Despite their impressive fluency, LLMs are prone to \emph{hallucination} - the generation of confident yet factually incorrect or unsupported content~\cite{huang_hallu_survey, sahoo_hallu_survey}. This undermines trust and remains a critical barrier in the adoption of LLMs in safety-critical applications. A growing body of work has sought to address this~\cite{selfcheckgpt, llmcheck, fadeeva2024fact, farquhar2024detecting, azaria-mitchell-2023-internal, zhou2025hademif, bazarova2025hallucinationdetectionllmstopological, orgad_llms_2025, binkowski2025hallucinationdetectionllmsusing}, yet most of the existing methods share a common limitation: they primarily operate on the final output text or shallow token-level statistics. They inspect the result of the model's reasoning and not the reasoning pathway.

\begin{figure}[t]
    \centering
    \includegraphics[width=\columnwidth]{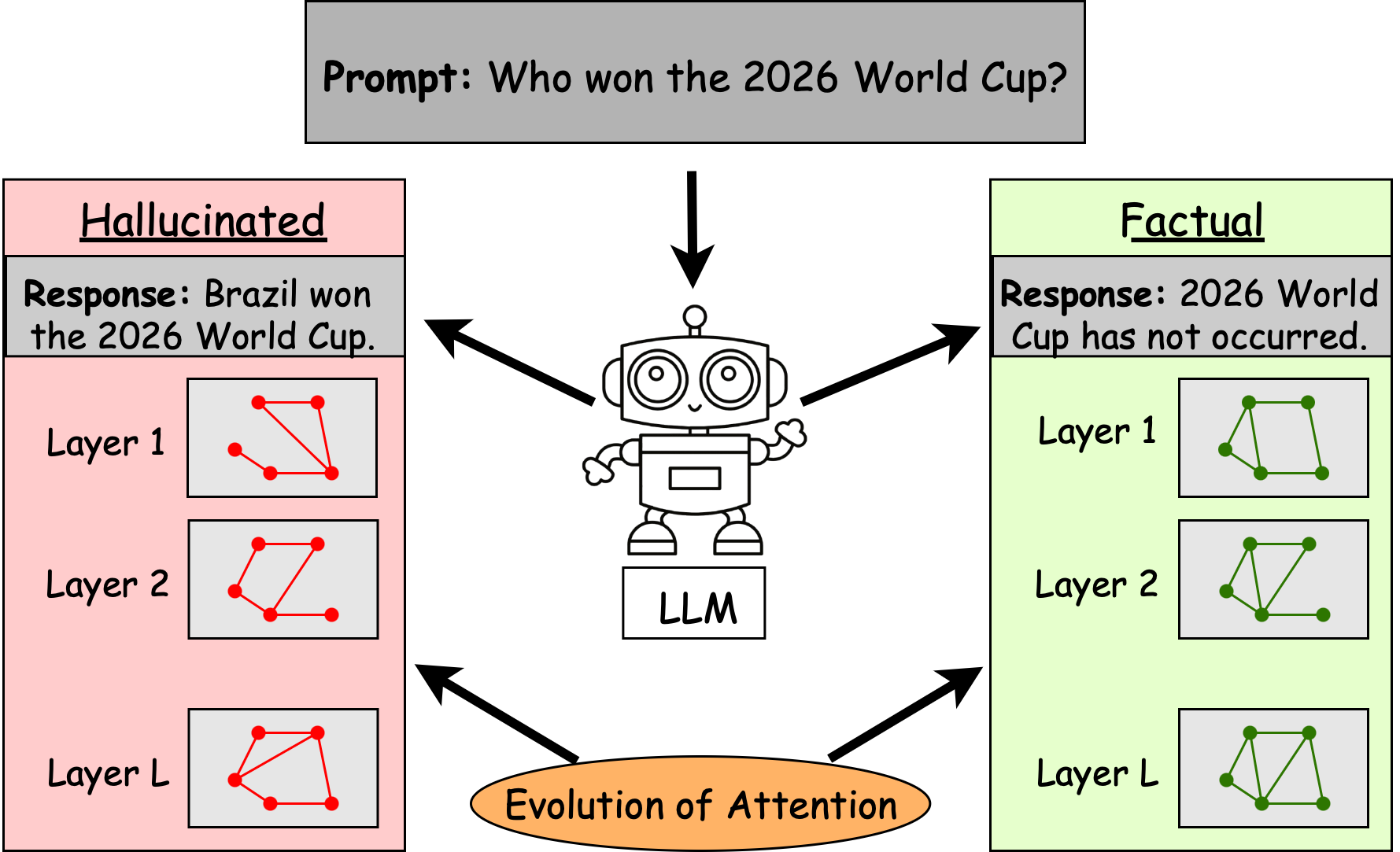}
    \caption{%The figure shows attention evolution patterns in factual versus hallucinated LLM responses. 
    Attention matrices modeled as graphs show distinct topological patterns as they evolve through a model's layers. We leverage zigzag persistence in topological data analysis to quantify these evolving attention structures for hallucination detection.}
    \label{fig:evoln_attn_hallu_detecn}
\end{figure}

To address this, we shift the focus from \emph{what} the model generates to \emph{how} it arrives at its conclusion. We hypothesize that a faithful generation relies on coherent flow of information, where tokens consistently attend to the relevant evidence across layers. Contrastingly, hallucination may arise when this flow breaks down and attention patterns diverge towards spurious contexts, leading to fabricated answers. Capturing these dynamics requires tools that can characterize how the structure of attention evolves across the model's layers.

Characterizing this evolving structure is a non-trivial task. While aggregate metrics of the attention matrix, such as eigenvalues~\cite{binkowski2025hallucinationdetectionllmsusing} or the determinant~\cite{llmcheck}, provide valuable insights, they do not capture higher-order structural properties. They cannot, for example, describe how distinct groups of tokens form conceptual clusters or how reasoning loops are formed. Topological Data Analysis (TDA) provides a principled way to bridge this gap.
% This is precisely the gap that Topological Data Analysis (TDA) addresses. 

% \begin{figure}[t]
%     \centering
%     \includegraphics[width=0.5\linewidth]{ARR_Oct_25/figs/evolving_attn_hallu_detecn.png}
%     \caption{Caption}
%     \label{fig:placeholder}
% \end{figure}

TDA~\cite{edelsbrunner2010computational, dey_wang_2022_book} provides a mathematical language to describe the `shape' of this evolving structure. Persistent Homology, a flagship concept of TDA, is a method for identifying the most robust structural features of a system by analyzing it at all scales simultaneously. This is achieved through a filtration, a process analogous to gradually lowering a threshold on attention weights to see which conceptual clusters and reasoning loops are fundamental (i.e., they persist for a long time) versus those that are ephemeral artifacts of noise. However, standard persistent homology is limited to analyzing systems that only grow, following a nested sequence ($G_0 \subseteq G_1 \subseteq \hdots $). 

This model is insufficient for capturing evolution of attention, where the structure is not merely augmented but is completely transformed between layers, with connections being both formed and broken. To capture this complex evolution, we leverage \emph{zigzag persistence}~\cite{carlsson2010zigzag,maria2016computing,carlsson2019parametrized,fzz}. Zigzag persistence is an extension of standard persistence designed to track topological changes through a series of inclusions (additions) and deletions (subtractions). Thus, viewing through the lens of zigzag persistence enables us to move beyond quantifying individual connections, to characterizing the topology of the evolving attention matrices. 

\begin{figure*}
    \centering    \includegraphics[width=1.9\columnwidth]{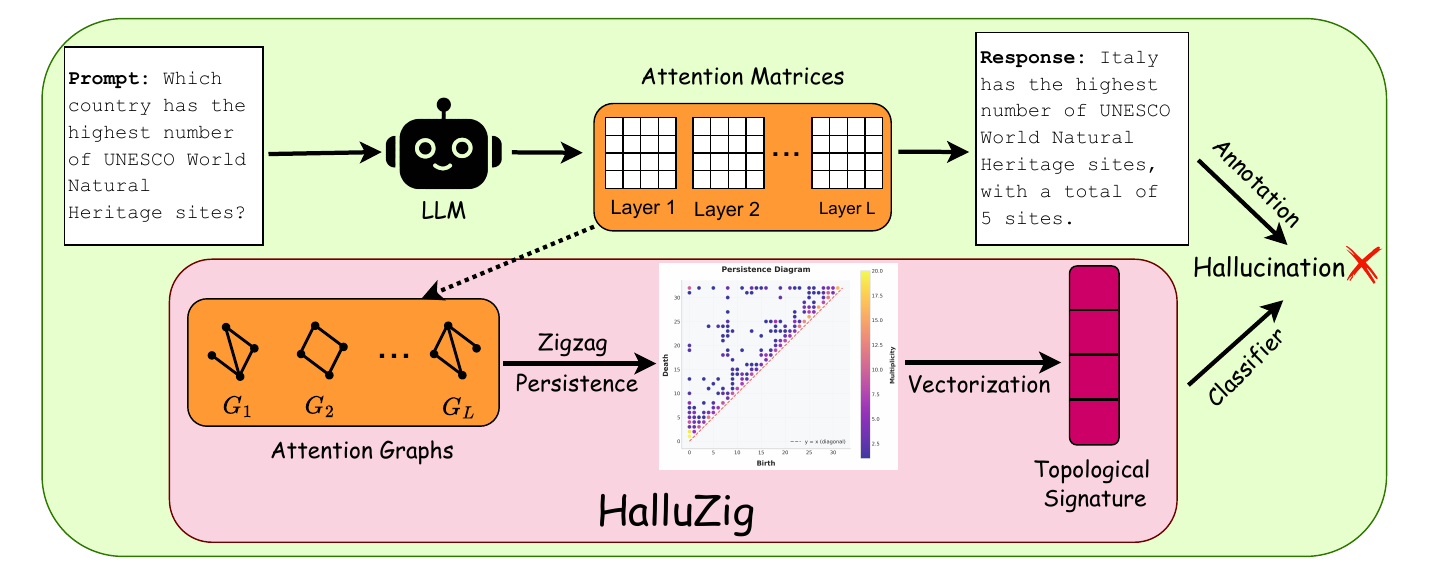}
    \caption{\textbf{The \topoframework{} framework: capturing attention evolution for hallucination detection.} We model the layer-wise attention matrices from an LLM as a sequence of attention graphs $(G_1, \hdots , G_L)$. Zigzag persistence is applied to this sequence to capture the evolution of topological features resulting in a Persistence Diagram. The Persistence Diagram is vectorized into a topological signature, which is used by a classifier to detect hallucinations.}
    \label{fig:halluzig_pipeline}
\end{figure*}

In this paper, we introduce \topoframework{} - a novel framework that captures the \emph{evolution of attention} in an LLM using \emph{zigzag persistence}. We model the attention matrix at each layer as a graph (\emph{attention graph)} and connect successive graphs through a \emph{zigzag filtration} ~\cite[Chapter 4]{dey_wang_2022_book}. By computing the zigzag persistence of this filtration, we obtain a topological summary quantifying births and deaths of connected components and cycles as information propagates through the model. This topological summary is distilled into a numerical vector to give a \emph{topological signature} using established methods~\cite{adams2017persistence, ATIENZA2020107509}.  Our core hypothesis is that factual and hallucinated responses leave distinct topological signatures in the evolution of the model's attention. To evaluate this hypothesis, we experiment with diverse datasets, which include human annotated datasets and generic QA based datasets which we annotate using \texttt{LLM-as-a-judge} paradigm~\cite{zheng_judging_2023}. Our experiments confirm the hypothesis, showing that the topological signatures can distinguish between factual and hallucinated responses. Further, we demonstrate the applicability of \topoframework{} for hallucination detection using partial network depth. We show that it can reliably detect hallucinations by analyzing only the first 70\% of the model's layers with minimal degradation in accuracy compared to a full-model analysis. Furthermore, we demonstrate that these topological signatures are not model-specific, exhibiting remarkable zero-shot generalization when transferred between different LLM architectures.

In summary, our main contributions are as follows: (1) We propose a new framework for hallucination detection by modeling the evolution of attention through an LLM. (2) To the best of our knowledge, this is the first application of zigzag persistence for capturing the layer-to-layer structural transformations of attention graphs for this task. (3) We demonstrate that \topoframework{} outperforms strong baselines on multiple datasets annotated for hallucination detection. (4) We empirically demonstrate that \topoframework{} achieves near-maximum performance when restricted to topological signatures from the first 70\% of model layers.

\section{Related Work}
\label{sec:related_work}

Research in hallucination detection has attracted a lot of attention in the recent times~\cite{huang_hallu_survey, zhang_enhancing_2023, wang_hidden_2024} can be broadly classified into \emph{black-box} methods and \emph{white-box} methods.

\paragraph{Black-box methods.} These methods operate only on the model's final text output. Consistency based techniques such as~\cite{selfcheckgpt, chen_inside_2024, kuhn2023semantic, qiusemantic, nikitin2024kernel} evaluate agreement among multiple generations. These methods rely on multiple model runs which imposes significant computational overhead. Surface-level confidence measures, such as perplexity, logit entropy, or predictive uncertainty~\cite{fadeeva2024fact, malinin2021uncertainty} provide lightweight alternatives but with limited discrimating power, as they do not consider the model's internal reasoning process.

\paragraph{White-box methods.} These methods leverage internal representations such as hidden states, attention maps, or logits. One of the early works in this direction was~\cite{azaria-mitchell-2023-internal} which had a linear probe into these states to determine factuality. Subsequent studies such as~\cite{farquhar2024detecting, chen_inside_2024, ch-wang-etal-2024-androids} quantified internal uncertainties by comparing hidden states across multiple generations. \cite{du_haloscope_2024} reduced annotation requirements while~\cite{kossen_semantic_2024} learned to approximate expensive self-consistency scores. A more recent and less explored white-box direction involves analyzing \emph{attention maps}. \newcite{chuang_lookback_2024} introduced lookback ratio which measures how strongly a model attends to relevant input tokens when generating context-dependent answers. \newcite{llmcheck} introduced simple attention statistics to flag a response as hallucinated in an unsupervised manner. \newcite{binkowski2025hallucinationdetectionllmsusing} use the eigen values of the attention matrix and the eigen values of the Laplacian of the attention matrix (modelled as a graph) to classify whether a response is hallucinated. \newcite{bazarova2025hallucinationdetectionllmstopological} introduce a topology-based hallucination detection technique which leverages a topological divergence metric between the prompt and the response subgraphs. These techniques typically exploit only local or scalar attention features, overlooking the rich, global structure of the evolution of attention graph that \topoframework{} aims to capture.

\section{Method}
\label{sec:method}

\subsection{Attention Mechanism}
\label{subsec:method:attention}
The \emph{self-attention}~\cite{vaswani2017attention} is the core component of the transformer architecture, which allows the LLM to dynamically weigh the importance of different tokens in a sequence when producing a representation for a given token.

At a high level, self-attention operates using three learned vector representations for each token: (1) Query($Q$), (2) Key($K$) and (3) Value($V$). Given a generated sequence of tokens $S = \{t_1, \hdots, t_T \}$, let $X \in \mathbb{R}^{T\times d}$ denote the matrix of $T$ tokens, each having a $d$-dimensional representation. Let $W_Q, W_K, W_V \in \mathbb{R}^{d \times d}$ denote the trainable projection matrices. Then, the three vector representations $Q, W$ and $V$ are $XW_Q, XW_K$ and $XW_V$ respectively. Note that $Q, K, V \in \mathbb{R}^{T \times d}$. The attention mechanism is defined as follows:
\begin{equation*}
    Attn(Q,K,V) = \text{softmax} \left (\frac{QK^T}{\sqrt{d}} \right ) V, 
\end{equation*}
where $d$ denotes the dimension of the token embedding. The matrix $A = \text{softmax} \left (\frac{QK^T}{\sqrt{d}} \right )$ is called the \emph{attention matrix}.

Modern LLMs use multi-head attention, which performs this process multiple times in parallel with different, learned projections for $Q, K$ and $V$. We denote the attention matrix at head $h$ of layer $l$ as $A^{(l,h)}$.

\subsection{Attention Graph Construction}
\label{subsec:method:attn_graph_cons}
\begin{figure}[htbp]
    \centering
    \begin{subfigure}[b]{0.48\columnwidth}
        \centering
        \includegraphics[width=\columnwidth]{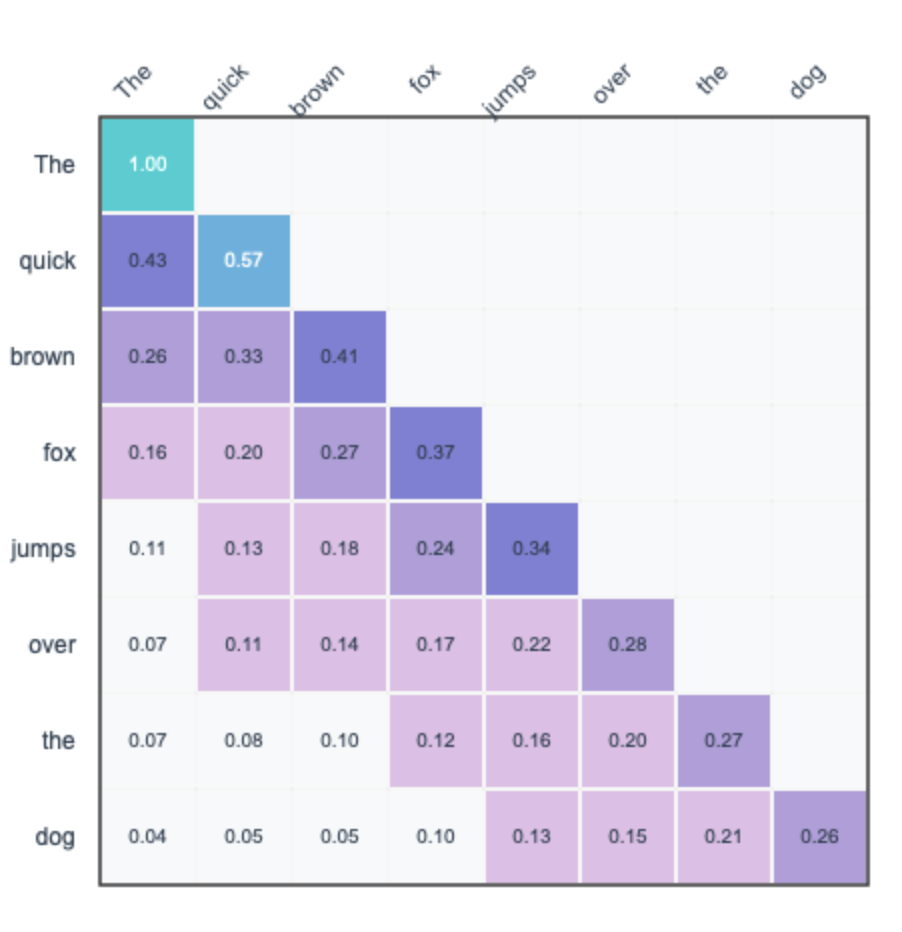}
        \caption{\textbf{A single-head attention matrix from a decoder-only model.} The matrix is lower-triangular due to the causal mask, which prevents tokens from attending to future positions in the sequence.}
        \label{fig:attn_mat}
    \end{subfigure}
    \hfill
    \begin{subfigure}[b]{0.48\columnwidth}
        \centering
        \includegraphics[width=\columnwidth]{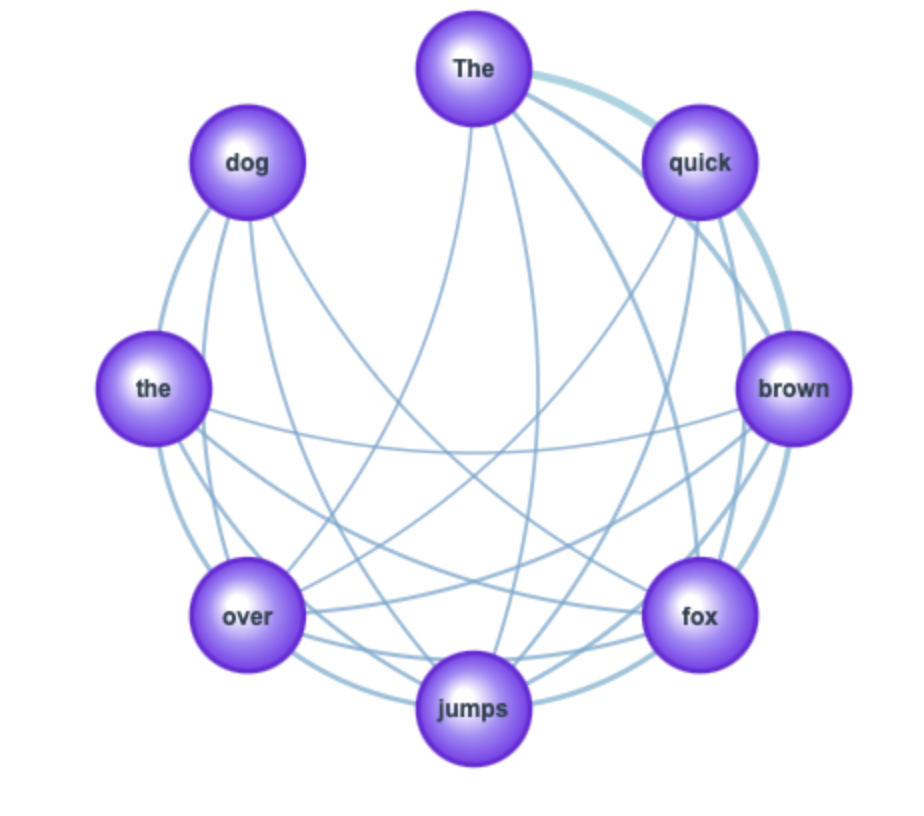}
        \caption{\textbf{An attention graph derived from the matrix in Figure~\ref{fig:attn_mat}.} The nodes represent tokens, while the edges visualize the of attention. The thickness of each edge corresponds to its attention weight, highlighting the key inter-token relationships that form the basis for our topological analysis.}
        \label{fig:attn_graph}
    \end{subfigure}
    \caption{\textbf{Visualizing the Attention Mechanism.} (a) A causally masked attention matrix. (b) The corresponding attention graph where nodes are tokens and thick edges represent high-attention links. This structural representation is the input to our topological pipeline.}
    \label{fig:attn_graph_mat}
\end{figure}

 For each layer $l \in \{1, 2, \hdots, L \}$, we average the attention matrices across all heads to get a mean attention matrix $A^{(l)} \in \mathbb{R}^{T \times T}$. We model this as a weighted, graph $G_l = (V, E_l, w_l)$, where $V$ is the set of tokens, and $w_l(t_i, t_j)$ is the attention weight from token $t_i$ to $t_j$. We choose the top $k$ percentile of attention weights to form the edges of the graph $G_l$ in order to focus on the most significant structural connections within the layer, ensuring our analysis is robust to the noise from low-value attention weights. We refer to these graphs as \emph{attention graphs}. This gives us a sequence of graphs, which captures the state of information flow at a specific depth in the model. Refer to Figure~\ref{fig:attn_graph_mat} for an illustration.

\subsection{Topological Signatures}
\label{subsec:method:top_prelim}

To analyze the structure of attention dynamics, we employ tools from TDA. TDA provides a mathematical framework for characterizing the "shape" of complex data. Here, we introduce the core concepts, building intuition from the familiar structure of an attention graph.

\begin{figure}
    \centering
    \includegraphics[width=\columnwidth]{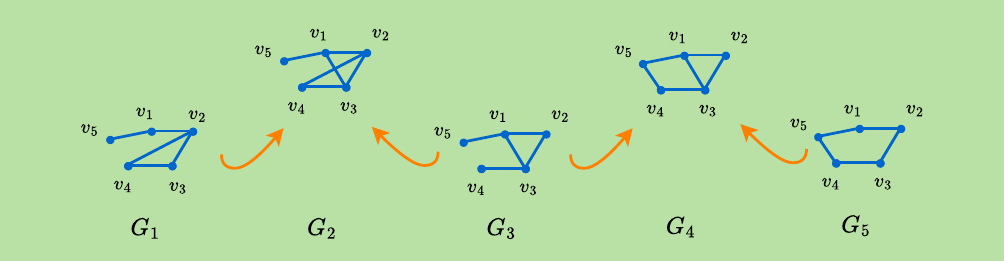}
    \caption{\textbf{Zigzag filtration.} The figure shows a zigzag filtration where $G_2 = G_1 \cup G_3$ and $G_4 = G_3 \cup G_5$.}
    \label{fig:zz_filt}
\end{figure}

Our starting point is an attention graph, where tokens are vertices and attention scores are weighted edges. TDA provides a formal language to describe the structure of such graphs that keep on changing.
\cancel{
through the concept of a \emph{simplicial complex}. Given a finite vertex set $V$, a simplicial complex $K=K(V)$ is a collection of subsets of $V$ such that if a subset $\sigma\subseteq V$ is in $K$, then all subsets $\tau\subset \sigma$ are also in $K$. Each element in $K$ with cardinality $k+1$ is called a $k$-simplex or simply a simplex. A graph is a simplicial complex where $V$ is the vertex set of the graph and $K$ consists of edges and vertices of the graph, where each vertex is a $0$-simplex and each edge is a $1$-simplex. By viewing our attention graph as a simplicial complex, we gain access to algebraic machinery for quantifying its structure.
}
A \emph{filtration} $\mathcal{F}$ of graphs is a nested sequence of 
graphs indexed by natural numbers/integers 
\begin{equation*}
    \mathcal{F}: G_0 \subseteq G_1 \subseteq G_2 \hdots \subseteq G_n.
\end{equation*}
Notice that the inclusions here are all in the forward direction, i.e., the sequence is a non-decreasing sequence of graphs. Now, if some of the inclusions were to be reversed, we get a \emph{zigzag filtration} $\cZ$ 
\begin{equation*}
    \cZ : G_0 \subseteq G_1 \supseteq G_2 \subseteq \hdots \supseteq G_n.
\end{equation*}
Refer to Figure~\ref{fig:zz_filt}. Unlike a standard filtration, which only allows the addition of new vertices/edges, a zigzag filtration also allows their removals. Attention values between tokens, as a sentence passes through various layers, keep evolving. To capture the evolution of attention graphs across layers, it is crucial to allow the deletion of edges. This is because two tokens that exhibit strong mutual attention in layer $l_i$ may no longer do so in layer $l_j$, requiring the corresponding edge to be removed from the attention graph at that layer. 

We obtain a zigzag persistence module $M_\cZ$ from such a zigzag filtration of graphs by computing the homology groups in dimensions $0$ (\# of connected components) and
$1$ (\# of independent cycles) of each graph $G_i$ appearing in the filtration. In other words, for each stage $G_i$, we record the $p$-dimensional topological features (such as connected components for $p=0$, loops for $p=1$). These homology groups are then connected by linear maps that follow the direction of the inclusion arrows in $\cZ$.
\begin{equation*}
    M_{\cZ}: H_p(G_0) \rightarrow H_p(G_1) \leftarrow  \hdots \leftarrow H_p(G_n)
\end{equation*}
 The persistence module $M_{\cZ}$ thus encodes how topological features appear, disappear, or reappear as the graphs evolve. We refer the reader to Appendix~\ref{app:defns} for formal definitions of (simplicial) homology groups. 

A key result in zigzag persistence is that any zigzag persistence module $M_\cZ$ decomposes into a direct sum of indecomposable interval modules~\cite{carlsson2010zigzag}. Each interval module corresponds to a contiguous range of indices in the zigzag filtration over which a homology class exists. Intuitively, each interval corresponds to the ``lifetime'' of an individual topological feature. This decomposition is unique up to reordering. This collection of intervals provides a complete and concise summary of the filtration's topological dynamics and can be visualized in two equivalent ways: 
\begin{itemize}
    \item Barcodes: A collection of horizontal line segments, where the length and the position of each line segment represents the lifetime of a feature.
    \item Persistence Diagrams: A 2D scatter plot of (birth, death) coordinates for each feature.
\end{itemize}

\begin{figure}[htbp]
    \centering
    \begin{subfigure}[b]{0.48\columnwidth}
        \centering
        \includegraphics[width=\columnwidth]{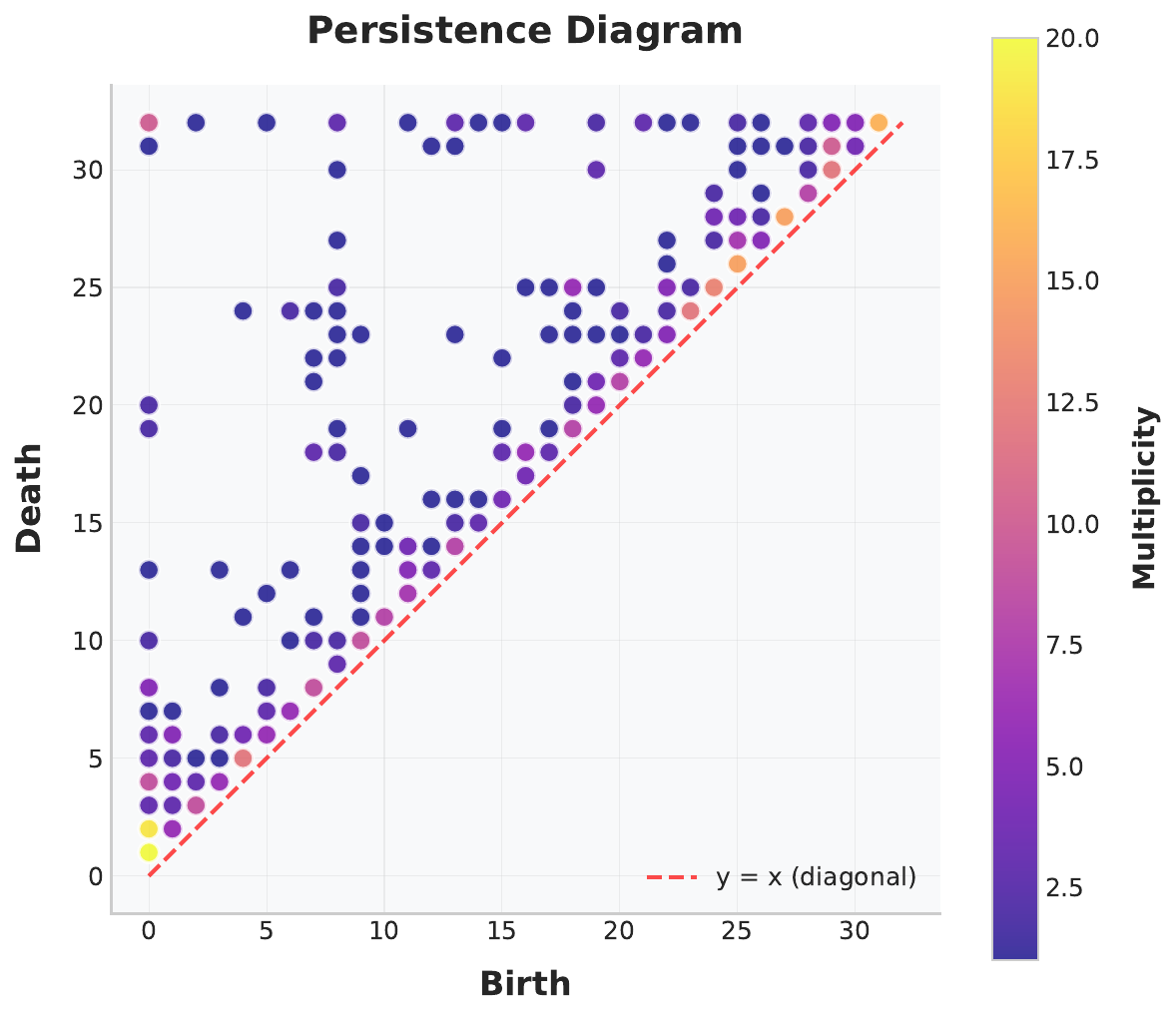}
        \caption{\textbf{Persistence Diagram.} Each point $(b, d)$ corresponds to a topological feature born at filtration value $b$ and dying at $d$. The vertical distance from the diagonal $y=x$, known as the feature's persistence.}
        \label{fig:pers_diag}
    \end{subfigure}
    \hfill
    \begin{subfigure}[b]{0.48\columnwidth}
        \centering
        \includegraphics[width=\columnwidth]{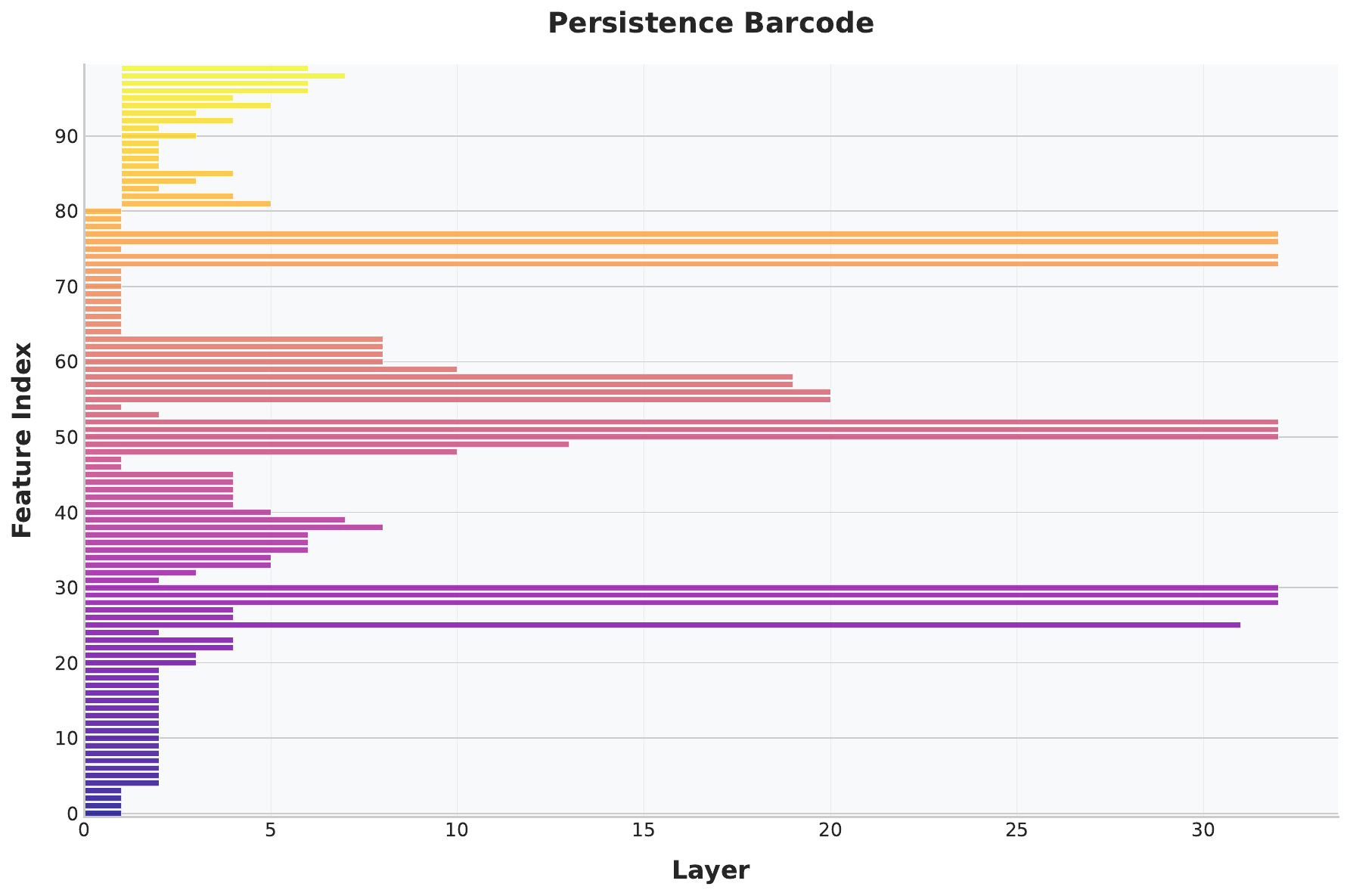}
        \caption{\textbf{Persistence Barcode.} Each horizontal bar represents a single topological feature, starting at its birth layer (x-axis) and ending at its death layer. The length of the bar directly visualizes the feature's persistence. For readability, this plot displays the first 100 bars, sorted vertically by birth time.}
        \label{fig:pers_barcode}
    \end{subfigure}
    \caption{\textbf{Two Views of a Topological Summary.} (a) The 2D persistence diagram plots each feature's birth vs. death layers. (b) The 1D barcode represents each feature's lifetime as a horizontal bar. }
    \label{fig:pers_diag_barcode}
\end{figure}

\subsection{Featurization and Classification}
\label{subsec:method:vectorization}
A persistence diagram is not a suitable input for most machine learning models. We need to transform each persistence diagram into a vector, which would be representative of the topological information captured in the persistence diagram. This vector can be used as input to a machine learning classifier. There are multiple known vectorization techniques such as persistence images~\cite{adams2017persistence}, persistence entropy~\cite{ATIENZA2020107509}, persistence landscapes~\cite{bubenik2015perslandscapes}, Betti curves~\cite{betti_curves}.

\subsection{Zigzag Persistence of Attention Dynamics}
\label{subsec:method:zz_pers}
To analyze the information flow between layers, we focus on consecutive pairs of graphs $G_l$ and $G_{l+1}$. We construct a short zigzag filtration between these two graphs as follows: 
\begin{equation*}
    G_l \hookrightarrow G_l \cup G_{l+1} \hookleftarrow G_{l+1}.
\end{equation*}
Here, $G_l \cup G_{l+1}$ is an intermediate graph containing the union of vertices and edges in $G_l$ and $G_{l+1}$. Since the vertex set is the same for graphs from each layer, $G_l \cup G_{l+1}$ is the union of the edges of $G_l$ and $G_{l+1}$. Refer to Figure~\ref{fig:zz_filt} for an illustration. The weight of the edge in $G_l \cup G_{l+1}$ is taken as the maximum of the weights of the edge in $G_l$ and $G_{l+1}$ if it exists in both graphs. This explicitly models how the structure of $G_l$ is transformed to produce the structure of $G_{l+1}$. The pairwise connections weave together all graphs from layer 1 to layer L to model the full sequence of attention dynamics.

In our case, the vertex set of the graphs do not change and thus
we focus on the evolution of loops (representing cyclic attention on tokens).
Algebraically, we track the evolution of the number of independent cycles or loops in $G_l$, which is the rank of the $1$-dimensional homology group $H_1(G_l)$. 
% Our reasoning is that cycles represent recurrent patterns of information flow, for instance, where token A attends to B, B to C, and C back to A. 
Such loops can be interpreted in two ways: as stable, resonant structures that reinforce a coherent semantic concept, or as flawed, circular reasoning pathways. We hypothesize that these two cases can be distinguished by their topological persistence. Cycles in factual statements are likely to be formed by strong, stable attention weights that persist across multiple layers, representing the successful consolidation of a concept. Conversely, we posit that hallucinations may manifest as numerous, short-lived, and structurally unstable cycles, indicative of spurious thought loops. By analyzing the 1-dimensional persistence diagrams, we aim to capture these signatures of flawed reasoning that are invisible to simpler connectivity measures.

In this paper, we use three vectorization schemes to vectorize the zigzag persistence diagrams and capture the topological information present in the barcodes. Different vectorization techniques present different representations of the underlying topological information: 
% \begin{itemize}

    \textbf{(1) Persistence Images (\texttt{PersImg})~\cite{adams2017persistence}:} PersImg treats the persistence diagram as a 2D distribution. It projects each persistence point $(b,d)$ onto a grid using a Gaussian kernel, effectively creating a heatmap or "image" that captures the geometric density and spatial relationships of topological features.
    
    \textbf{(2) Persistence Entropy (\texttt{PersEntropy})} ~\cite{ATIENZA2020107509}: PersEntropy provides a statistical summary of the barcode. It calculates the Shannon entropy~\cite{shannon} of the distribution of feature lifetimes (the lengths of the bars or equivalently, $(d-b)$ for a point $(b,d)$ in the persistence diagram). A single entropy value quantifies the overall complexity of the topological signature, with higher values indicating more uniform persistence across features.
    
    \textbf{(3) Betti Curve~\cite{betti_curves}}: This method produces a 1D vector by plotting the Betti number, the count of currently active bars, as a function of the filtration value. The resulting curve, which tracks the feature counts, is then sampled at discrete points to form the feature vector.
% \end{itemize}

For simplicity, we refer to our overall approach as \topoframework{}, regardless of the specific vectorization scheme employed.

\section{Experiments}
\label{sec:exp}

\begin{table*}[!htbp]
    \centering
    \small
    % \resizebox{0.8 \textwidth}{!}{
    \begin{tabular}{c|c|ccc|c}
    \toprule
         Model&  Measure&AUCROC& Accuracy&TPR @ 5\% FPR &F1 Score\\
         \midrule
         &  Self-Prompt&50.30&  50.30& -&66.53\\
 & FAVA Model&53.29&53.29& -&43.88\\
 &SelfCheckGPT-Prompt&50.08&54.19&  -&67.24\\
 & INSIDE&59.03& 57.98& 13.17&39.66\\
\texttt{Llama-2-7b} & LLM-Check (Attn Score)&72.34& 67.96& 14.97&69.27\\
 & \texttt{PersImg}& \textbf{82.09}& \textbf{75.00}& \textbf{26.79}&\textbf{80.67}\\
 & \texttt{PersEntropy}& 75.67& 72.82& 21.43&78.26\\
 & \texttt{Betti Curve}& 74.75& 68.47& 23.21& 77.86\\
 \midrule
 & LLM-Check (Attn Score) & 68.19 & 65.87 & 15.57 & 70.53 \\
 & \texttt{PersImg}& \textbf{82.64} &\textbf{73.91} &\textbf{46.43} &\textbf{80.33}\\
  \texttt{Llama-3-8b} & \texttt{PersEntropy}& 74.09 &71.73 &19.29 &78.69 \\ 
  & \texttt{Betti Curve}&74.33 & 70.65& 21.43&77.69 \\
\midrule
& LLM-Check (Attn Score) & 71.69 & 66.47 & 24.55 & 62.00 \\
 & \texttt{PersImg}& \textbf{83.28} &\textbf{77.17} & \textbf{35.71}&\textbf{82.35}\\
  \texttt{Vicuna-7b}& \texttt{PersEntropy}&75.47 &68.47 &17.86 &76.03\\
 & \texttt{Betti Curve}&76.71 &70.65 &24.59 &80.29\\
  \bottomrule
    \end{tabular}
    % }
    \caption{\textbf{Hallucination detection results on the FAVA Annotated Dataset.} The LLM-Check (Attn Score), Self-Prompt, FAVA Model, SelfCheckGPT-Prompt and INSIDE results are according to the numbers in ~\newcite{llmcheck}.}
    \label{tab:fava_annot}
    \vspace{-0.2cm}
\end{table*}
% In this section we detail the experimental setup and present the empirical results of our study.

\subsection{Experimental Setup}
\label{subsec:exp:exp_setup}
\paragraph{Datasets.} To ensure comprehensive evaluation, we test \topoframework{} on a diverse suite of benchmarks covering different domains and annotation styles:
% \begin{itemize}

\textit{Generative Benchmarks:} We use two benchmarks with explicit human provided hallucination labels. The FAVA Annotated Dataset~\cite{mishra2024finegrained} provides passage-level binary labels for Wikipedia abstract generation. The RAGTruth Summarization Dataset~\cite{niu2024ragtruth} offers span-level annotations, which we normalize to passage-level task: a summary is considered if it contains one or more annotated spans.

\textit{QA-based Benchmarks:} We assess the performance on TruthfulQA~\cite{lin2021truthfulqa} and NQ-Open~\cite{kwiatkowski_natural_2019} datasets. As these datasets do not contain explicit hallucination labels, we employ the \texttt{LLM-as-a-judge} paradigm~\cite{zheng_judging_2023}. We use GPT-4o-mini~\cite{openai2024gpt4ocard}, a closed-source LLM, to automatically annotate whether a generated answer is a hallucination or not.
% \end{itemize}

\paragraph{Models and metrics.} For our experiments, we use a diverse set of open-source LLMs, including models from the Llama family (\texttt{Llama-2-7b, Llama-2-13b}~\cite{touvron2023llama}, \texttt{Llama-3-8b, Llama-3.1-8b, Llama-3.2-3b}~\cite{grattafiori2024llama3herdmodels}),  \texttt{Vicuna-7b}~\cite{vicuna2023} and \texttt{Mistral-7b}~\cite{jiang2023mistral7b} , all accessed via the Hugging Face transformers library \cite{wolf_transformers_2020}. Our topological pipeline is implemented using the FastZigzag library \cite{fzz} for persistence computation and the Gudhi library \cite{gudhi:PersistenceRepresentationsScikitlearnInterface} for vectorization. We employ a Random Forest Classifier for the final binary classification and evaluate its performance using Accuracy, F1-Score, AUC-ROC, and TPR@5\%FPR. For clarity in our main results, we report the performance of the best run across multiple random seeds. To ensure full reproducibility, we provide the mean and standard deviation for all experiments, along with a detailed list of all hyperparameters, in Appendix~\ref{app:exps}. Moreover, all the results reported in the paper are the result of a single LLM run of the respective models. The code is available at \href{https://github.com/TDA-Jyamiti/halluzig}{https://github.com/TDA-Jyamiti/halluzig}

\paragraph{Baselines.} Since our method leverages structural signals embedded in attention matrices, we evaluate our approach against baselines that capture analogous information. On generative benchmarks (FAVA Annotated Dataset and RAGTruth Summarization Dataset), we compare against the Attention Score metric from the LLM-Check~\cite{llmcheck}. To provide additional context on where \topoframework{} stands with respect to other uncertainty quantification methods, we add SelfPrompt~\cite{kadavath2022languagemodelsmostlyknow}, INSIDE~\cite{chen_inside_2024}, SelfCheckGPT-Prompt~\cite{selfcheckgpt} and FAVA Model~\cite{mishra2024finegrained} to the comparisoin.  For the QA benchmarks, we add LapEigVals~\cite{binkowski2025hallucinationdetectionllmsusing} to the comparison as it extracts spectral properties from attention matrices. 

% \vspace{-0.2cm}
\subsection{Main Results}
\label{subsec:exp:main_results}

\topoframework{} outperforms the baselines across models on most datasets. The key empirical findings are summarized below.

\begin{table}[!htbp]
  \centering
  % \small
  \resizebox{\columnwidth}{!}{
  \begin{tabular}{c|c|c|c|c|}
    \toprule
    \multirow{1}{*}{\textbf{Method}} & \multirow{1}{*}{\textbf{Metric}} & \multicolumn{1}{c|}{\textbf{White-box}} & \multicolumn{2}{c|}{\textbf{Black-box}}  \\
     \cmidrule(lr){3-5}
    & & Llama-2-7b & Llama-2-13b & Mistral-7b \\
    \midrule
    \multirow{4}{*}{Attention Score}
      & AUCROC        & 54.19 & 60.05 &  55.37\\
      & Accuracy      & 54.52 & 59.66 &  56.99\\
      & TPR @ 5\% FPR &  5.88 & 14.48 &  5.18\\
      & F1 Score      & 54.50 & 55.97 &  57.72\\
    \midrule
    \multirow{4}{*}{\texttt{PersImg}}
      & AUCROC     &  \textbf{73.37}  & \textbf{72.90} &  \textbf{74.45}\\
      & Accuracy      & \textbf{62.44} & \textbf{61.63} &  \textbf{63.45}\\
      & TPR @ 5\% FPR & 19.03 & \textbf{18.39} &  \textbf{26.22}\\
      & F1 Score      & \textbf{69.10} & \textbf{68.63} &  \textbf{70.84}\\
    \midrule
    \multirow{4}{*}{\texttt{PersEntropy}}
      & AUCROC        & 51.46 & 52.78 &  52.97\\
      & Accuracy      & 52.00 & 53.95 &  52.02\\
      & TPR @ 5\% FPR &  7.96 &  2.69 &  3.11\\
      & F1 Score      & 60.15 & 61.22 &  58.37\\
    \midrule
    \multirow{4}{*}{\texttt{Betti Curve}}
      & AUCROC        & 69.26 & 67.93 &  71.03\\
      & Accuracy      & 57.78 & 60.05 &  63.00\\
      & TPR @ 5\% FPR & \textbf{23.89} & 14.80 &  19.11\\
      & F1 Score      & 65.95 & 67.99 &  70.69\\
    \bottomrule
  \end{tabular}
  }
  \caption{\textbf{Hallucination detection performance on the RAGTruth Summarization Dataset.} We compare \topoframework{} (\texttt{PersImg, PersEntropy, Betti Curve}) with Attention Score from~\cite{llmcheck}. Results are reported for both a white-box setting (using the Llama-2-7b generator) and a black-box setting (using Llama-2-13b and Mistral-7b as substitutes).}
  \label{tab:ragtruth}
  \vspace{-0.2cm}
\end{table}

\begin{figure*}
    \centering
    \includegraphics[scale=0.9]{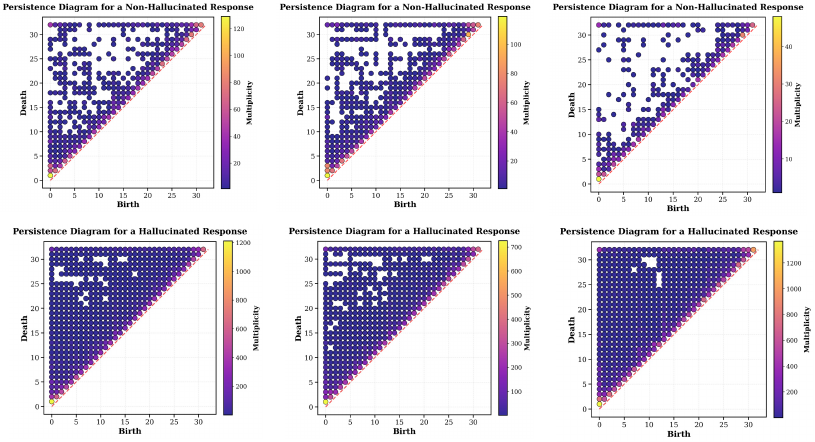}
    \caption{\textbf{Visualization of persistence diagrams for hallucinated versus non-hallucinated responses for FAVA dataset.} The top row depicts the persistence diagrams for three randomly selected non-hallucinated responses while the bottom row depicts the persistence diagrams for three randomly selected hallucinated responses from FAVA dataset. The persistence diagrams are colored by multiplicity, i.e., the multiplicity of each point in the diagram is depicted by its color. We can see that the persistence diagrams look visually different for hallucinated versus non-hallucinated responses which gets reflected in the \topoframework{} performance.}
    \label{fig:pers_diags_hallu_non_hallu}
\end{figure*}

\paragraph{Performance on Generative Benchmarks.} On the FAVA Annotated~\cite{mishra2024finegrained} and RAGTruth Summarization Datasets~\cite{niu2024ragtruth}, \topoframework{} achieves superior performance (Table~\ref{tab:fava_annot} and Table~\ref{tab:ragtruth}). A key takeaway is the robustness of the underlying topological signal: all three vectorization schemes (persistence images, Betti curves, and persistent entropy) yield competitive results, confirming that the structural information is the dominant feature. Furthermore, \topoframework{} excels in both black-box and white-box settings on RAGTruth Summarization dataset.  While performance varies by vectorization, both \texttt{PersImg} and \texttt{Betti Curve} significantly outperform the baseline in both scenarios. This demonstrates that the dynamic structural information captured by \topoframework{} is a robust signal for hallucination, even when direct model access is limited. Refer to Figure~\ref{fig:pers_diags_hallu_non_hallu} for an illustration.

\begin{table}[!htbp]
  \centering
  % \small
  \resizebox{\columnwidth}{!}{
  \begin{tabular}{c|c|cc}
    \toprule
    \textbf{LLM} & \textbf{Feature} & \textbf{NQOpen} & \textbf{TruthfulQA} \\
    \midrule
    \multirow{7}{*}{\texttt{Llama-3.1-8b}} 
      & AttentionScore & 0.556 & 0.541 \\ % remove
      & AttnEigvals    & 0.732 & 0.587 \\
      & LapEigvals     & \textbf{0.748} & 0.589 \\
 & \texttt{PersImg}& 0.730&\textbf{0.733}\\
 & \texttt{PersEntropy}&0.682 &0.664\\
 & \texttt{Betti Curve}&0.715 &0.684\\
 \midrule
    \addlinespace
    \multirow{7}{*}{\texttt{Llama-3.2-3b}} 
      & AttentionScore & 0.546 & 0.581 \\ % remove
      & AttnEigvals    & 0.694 & 0.535\\
      & LapEigvals     & 0.693 & 0.539 \\
 & \texttt{PersImg}& \textbf{0.712}& 0.641\\
 & \texttt{PersEntropy}&0.685 &0.656\\
 & \texttt{Betti Curve}& 0.688 &\textbf{0.667}\\
 \bottomrule
  \end{tabular}
  }
  \caption{\textbf{Hallucination detection performance on QA Benchmark.} Test AUROC scores for baseline and \topoframework{}. The results for AttentionScore, AttnEigVals and LapEigvals are based on the experiments we performed with their methods.}
  \label{tab:qa_based_detection}
\end{table}

\paragraph{Performance on QA Benchmarks.} \topoframework{}'s effectiveness extends to the QA domain (Table~\ref{tab:qa_based_detection}). \topoframework{} shows consistent improvement on both, TruthfulQA~\cite{lin2021truthfulqa} and NQ-Open~\cite{kwiatkowski_natural_2019} datasets, demonstrating that topological features generalize from controlled generative tasks to open-domain question answering.

We note that there is a discrepancy in the LapEigVal score reported in Table~\ref{tab:qa_based_detection} and that reported in their original paper~\cite{binkowski2025hallucinationdetectionllmsusing}. One of the main reasons is that using GPT-4o-mini as annotator produced a heavily class imbalanced dataset, (711/91/15) in our case versus (\textasciitilde{}500/\textasciitilde{}250/\textasciitilde{}80) reported in~\cite{binkowski2025hallucinationdetectionllmsusing}. Moreover, we used the temperature values as 0.7 for all our experiments while the results reported in~\cite{binkowski2025hallucinationdetectionllmsusing} are with 0.1 and 1.0. Our aim for these experiments was only to ensure a fair comparison between \topoframework{} and other baselines under similar labeling conditions.

We can see from the results that, generally, \texttt{PersImg} has a higher performance than \texttt{BettiCurve} and \texttt{PersEntropy}. Persistence image is a 2D heatmap (distribution) that captures the spatial relationship between birth and death layers in a persistence diagram. Betti Curve generally performs the second best which is a 1-D vector of counts of the number of features at different points over the filtration. Persistence Entropy is a single value summary of the persistence diagram. While still useful, quantifying the entire 2D plot (persistence diagram) by a single number inherently loses information, explaining its lower but competitive performance.

% \vspace{-0.4cm}
\section{Analysis and Discussion}
\label{sec:analysis_discussion}
% \vspace{-0.1cm}
The results demonstrate the efficacy of our method and also provide deeper insights into the nature of hallucinations and the potential for new model safety mechanisms. We analyze the implications of our key experimental findings.

% \vspace{-0.1cm}
\subsection{The Critical Role of Dynamics: Zigzag vs. Static TDA}
\label{subsec:analysis:zz_vs_static}
% \vspace{-0.1cm}

To validate our central hypothesis that the dynamics of attention evolution are more informative than static snapshots, we conducted an ablation study comparing our full model against a Static TDA baseline. This baseline computes standard persistent homology on each layer individually, deliberately ignoring the layer-to-layer transformations.

\begin{figure}
    \centering
    \includegraphics[width=\columnwidth]{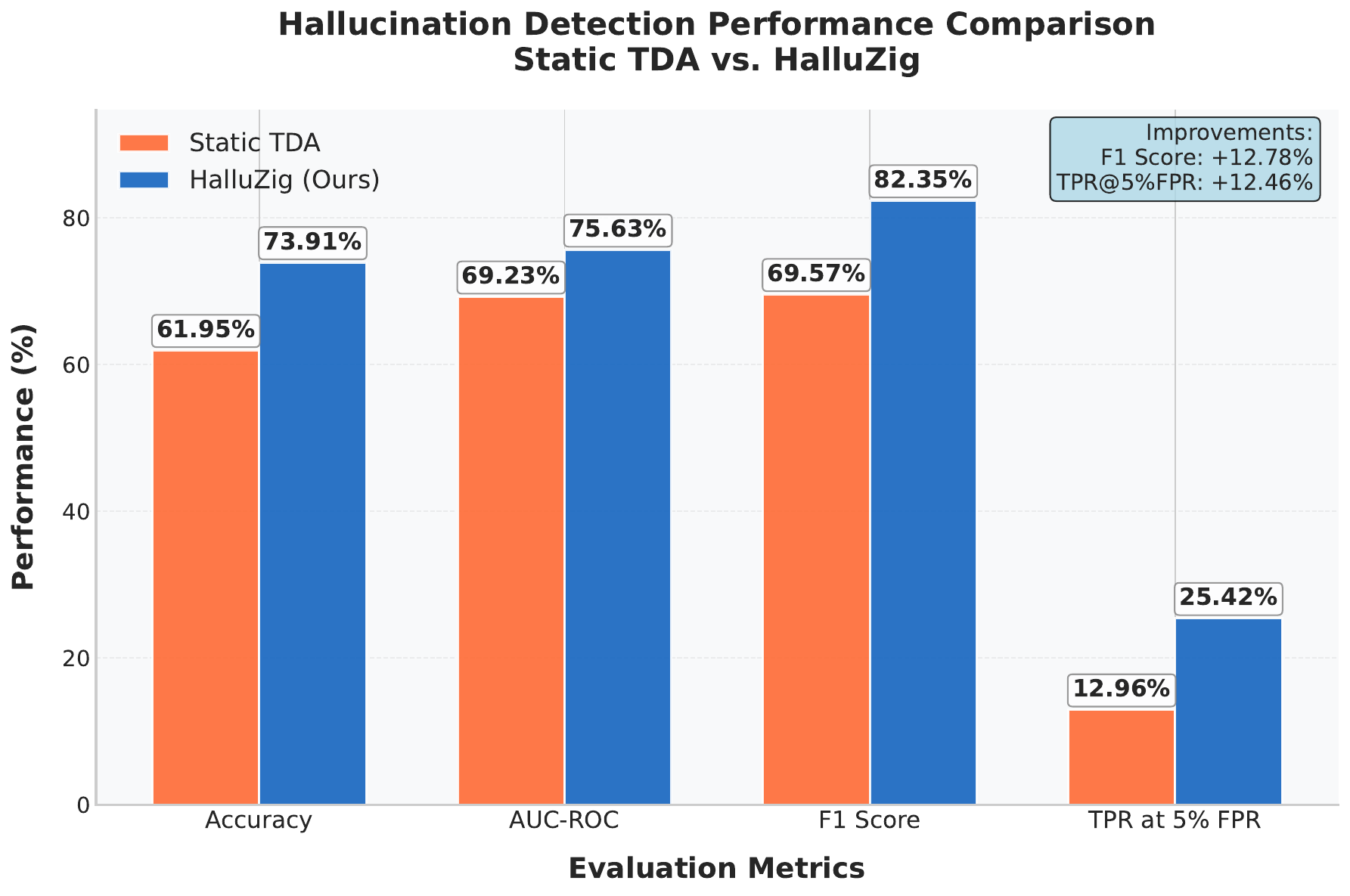}
    \caption{\textbf{Role of Dynamics: Comparing \topoframework{} against a baseline using static, layer-wise persistent homology.} Our full model demonstrates improved performance across all metrics, notably achieving a +12.78\% F1-score. The results underscore the importance of modeling the dynamic evolution of attention graphs over static analysis.}
    \label{fig:static_vs_zigzag}
    \vspace{-0.6cm}
\end{figure}

As shown in Figure~\ref{fig:static_vs_zigzag}, the Static TDA baseline underperforms \topoframework{} by a significant margin. This provides compelling evidence that a substantial amount of discriminative information is lost when inter-layer dynamics are disregarded. The key takeaway is that the crucial signal for hallucination lies not in the static topology of individual layers, but in the dynamics of their evolution, a property that zigzag persistence is equipped to capture, which supports our core hypothesis.

\subsection{Universality of Signatures: Cross-Model Generalization}
\label{subsec:analysis:cross_model_gen}
\begin{figure}
    \centering
    \includegraphics[width=\columnwidth]{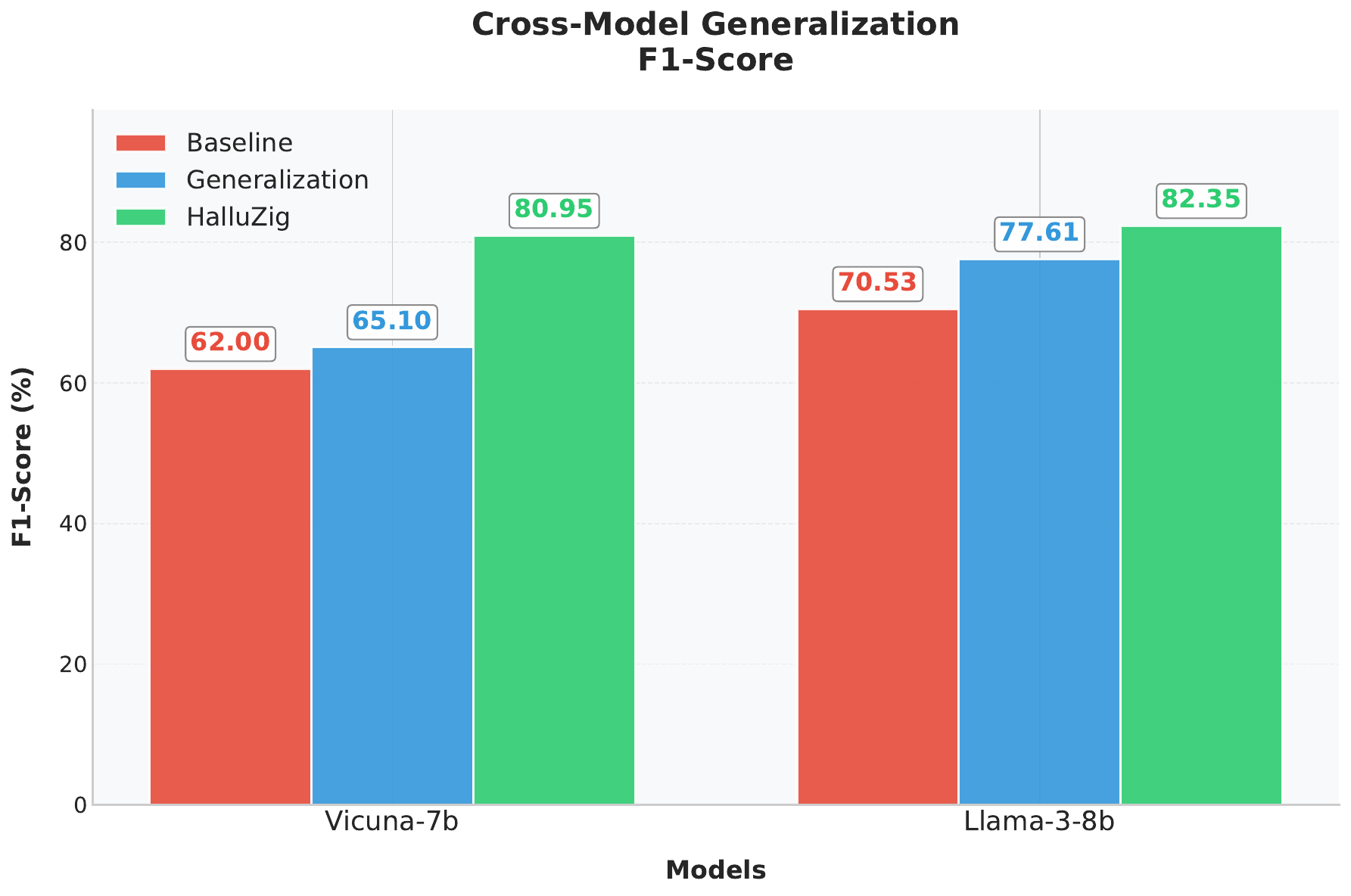}
    \caption{\textbf{Zero-Shot Cross-Model Generalization Performance.} The figure evaluates the ability of \topoframework{} to generalize across different LLM architectures without retraining. (Left) Performance of the classifier on \texttt{Llama-3-8b} trained exclusively on topological signatures extracted from \texttt{Vicuna-7b}. The generalization model outperforms the baseline (Attn-Score~\cite{llmcheck}) by 3\%. (Right) The reverse scenario, showing performance on \texttt{Vicuna-7b} of the classifier trained on \texttt{Llama-3-8b}. The generalization model surpasses the baseline by 7\% in this case. }
    \label{fig:cross_model_gen}
    % \vspace{-0.5cm}
\end{figure}

A critical question is whether the learned topological signatures are universal or model-specific. We investigated this via a zero-shot cross-model experiment, training a classifier on topological signatures extracted from \texttt{Llama-3-8b} and testing it on the topological signatures from \texttt{Vicuna-7b} and the vice-versa, without retraining. We use FAVA Annotated Dataset~\cite{mishra2024finegrained} for this analysis. The results reveal a remarkable degree of transferability. This result indicates that different LLMs may exhibit similar topological dynamics when hallucinating. 
% This suggests that the structural properties we capture are not just artifacts of a specific model's weights, but may be linked to more general properties of the attention mechanism's failure modes. 
\begin{figure}
    \centering
    \includegraphics[width=0.9\columnwidth]{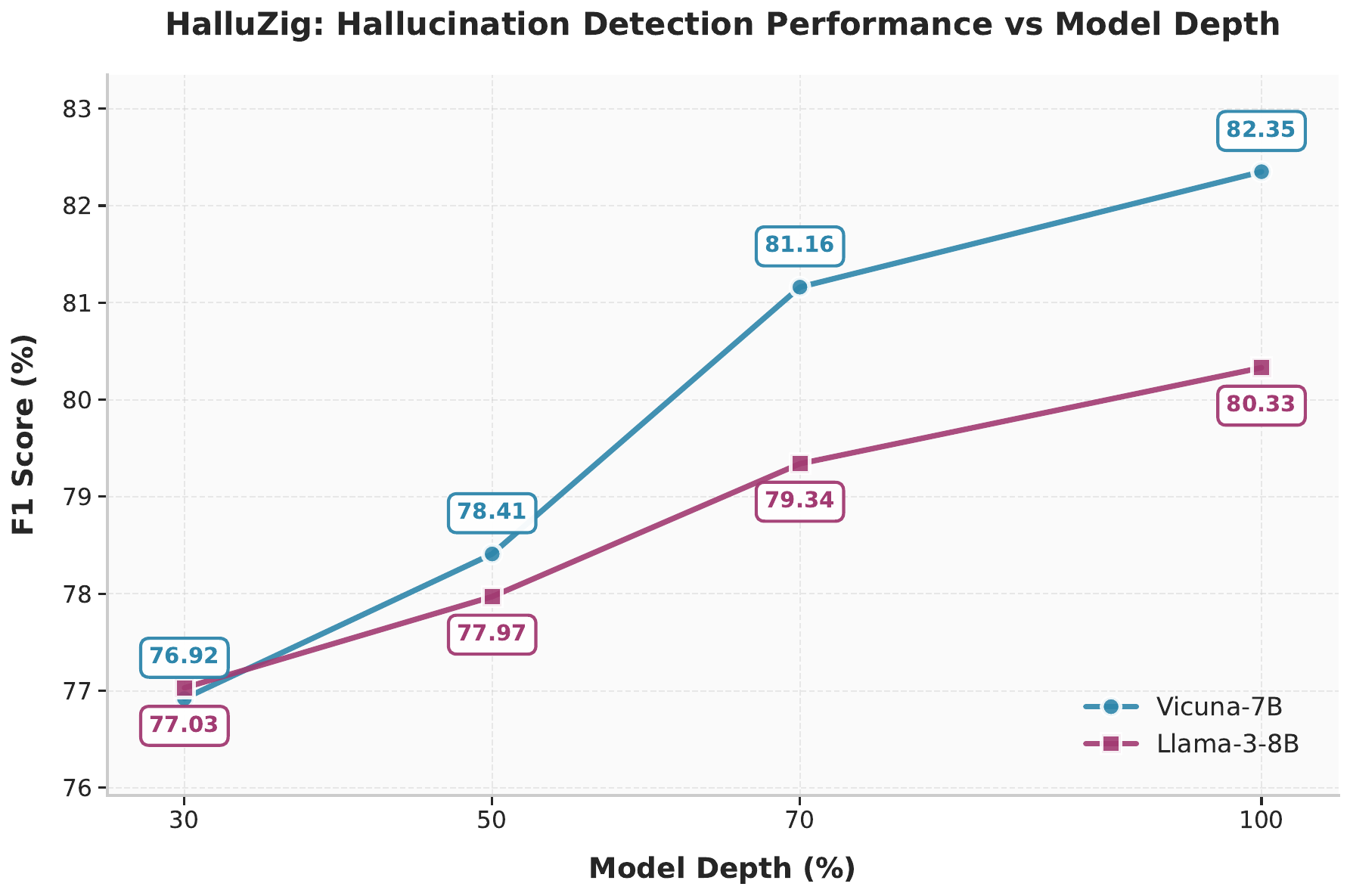}
    \caption{\textbf{Hallucination detection performance as a function of model depth.} The plot shows the F1-scores achieved when using topological features from an increasing percentage of the model's total layers. We observe that \topoframework{} performance achieves over 98\% of the final score by the 70\% depth mark. This indicates that the structural signatures of hallucination are formed in the middle layers.}
    \label{fig:early_detection}
    % \vspace{-0.5cm}
\end{figure}
\subsection{Practical Implications: Hallucination Detection from Partial Network Depth}

\label{subsec:analysis:early_detection}

Our final analysis investigates whether hallucinations can be detected reliably before the final layer. We computed topological features (\texttt{PersImg}) up to varying model depths of \texttt{Vicuna-7b} and \texttt{Llama-3-8b} on the FAVA Annotated dataset~\cite{mishra2024finegrained} and evaluated performance at each stage (Figure~\ref{fig:early_detection}). We observe from the performance curves that \topoframework{} achieves an F1-score nearly identical to that of the full-model analysis just at 70\% of the model's depth for both models.

This result suggests that hallucination may not be a last-minute failure but is encoded and stabilized relatively early in the model's reasoning pathway.

Together, these analyses reinforce that \topoframework{} is not only as a diagnostic tool but can also be a safety monitoring mechanism in LLMs.

%%%%%%%%%%%%%%%%%%%%%%%%%%%%%%%%%%%%%%%%%%%%%%%%%%%%%%%%%%%%%%%%%%%%%%%%%%

\section{Conclusion}
\vspace{-0.1cm}
In this paper, we introduced \topoframework{}, a novel approach for detecting hallucination by analyzing the topological information hidden in the evolution of attention in an LLM. By leveraging zigzag persistence, we demonstrated that the structural evolution of attention provides a powerful and robust signal for hallucination detection. Our experiments show that this method not only outperforms strong baselines but also exhibits remarkable cross-model generalization and enables reliable early detection. This work establishes the viability of structural interpretability, offering a new lens to understand and improve the trustworthiness of LLMs.

\section*{Acknowledgements}
This work is partially supported by NSF grants DMS-2301360 and CCF-2437030.

\section*{Limitations}
% \vspace{-0.2cm}
While our work establishes the viability of topological analysis of evolving attention for hallucination detection, we point out some limitations that can be in agenda for future research.

\subsection*{Computational Complexity}
The primary limitation of our current approach is the computational cost of computing persistent homology, particularly for long sequences (sentences) which result in large attention graphs. However, we note that this computational challenge is not unique to our framework, but is a well-known limitation within the applied TDA community.

\subsection*{Scope of Analysis}
Our framework focuses exclusively on the attention mechanism. However, other model components, particularly the MLP layers, are known to store and manipulate factual knowledge. Our approach is currently blind to procedural failures that may originate solely within these components. A more holistic approach could integrate topological signals from both attention and MLP activations.

\subsection*{Attention-Head Averaging}
A methodological choice of our framework is the mean-pooling of attention heads to create a single graph per layer. While this provides a stable, holistic view of information flow and ensures computational tractability, it is an important limitation. It is known that attention heads can specialize in distinct functions, and our averaging approach may dilute or obscure a strong, localized signal from a single ``rogue'' head whose aberrant behavior is the primary cause of a hallucination. Future work could pursue more fine-grained, head-specific topological analysis to gain deeper diagnostic insights, though this would entail a significant increase in computational cost.

\subsection*{Scope of Model Scale}
Our experimental validation is conducted on open-source LLMs with up to 13 billion parameters. This leads to the natural question of whether our findings on the topological dynamics of hallucination generalize to much larger, state-of-the-art foundation models (e.g., 70B+ parameters). We hypothesize that the observed structural patterns are a fundamental property of the Transformer architecture and will therefore apply to larger models. However, empirically verifying this scalability is a direction for future research.

\section*{Ethical Considerations}
% \vspace{-0.2cm}
To the best of our knowledge, we did not violate any ethical code while conducting the research work described in this paper. We report the technical details needed for reproducing the results and will release the code upon acceptance. All results are from a machine learning model and should be interpreted as such. The LLMs used to generate attention matrices for this paper are publicly available and are allowed for scientific research.

\bibliography{ref}

\appendix

% ===== APPENDIX A: TDA CONCEPTS =====
% \onecolumn
\section{Topological Data Analysis Preliminaries}
\label{app:defns}

This appendix provides the necessary background in topological data analysis, with particular emphasis on concepts relevant to zigzag persistence and its application to graphs.\\

\subsection{Simplicial Homology}

We begin by introducing simplicial complexes. Simplicial complexes are spaces built with smaller geometric objects (simplices), such as vertices, edges, filled triangles, and so on. More formally:

An \emph{abstract simplicial complex} $K$ is a family of non-empty subsets of an underlying finite set $V(K)$ that is closed under the operation of taking subsets. In other words, if $\sigma \in K$ and $\tau\subset \sigma$ then $\tau \in K$. Each such element $\sigma \in K$, is called a $p$-simplex if $|\sigma|=p+1$. A $\tau \subset\sigma$ is called a face of $\sigma$.

 A graph $G=(V,E)$ is easily seen as abstract simplicial complex, where the underlying set is the set of vertices $V=V(K)$, and the family of subsets $K$ is given by the union of set of vertices and the set of edges. Then the vertices correspond to the $0$-simplices and the set of edges to the $1$-simplices. 

For an abstract simplicial complex \( K \), define for each \( p \geq 0 \) a \emph{chain group} $C_p(K, \mathbb{Z}_2)$,
which is the vector space over $\mathbb{Z}_2$ generated by the \( p \)-simplices of \( K \). We usually omit the field $\mathbb{Z}_2$ in our notation. Elements of \( C_p(K) \) are called \emph{\( p \)-chains}, and they are formal sums of \( p \)-simplices with $\mathbb{Z}_2$ coefficients. One can connect $C_p(K)$ and $C_{p-1}(K)$ by a boundary operator $\partial_p : C_p(K) \to C_{p-1}(K)$, which maps each  \( p \)-simplex to the sum of its  $(p\!-\!1)$-dimensional faces. These boundary operators satisfy the key property $\partial_{p-1} \circ \partial_p = 0$.

Using these boundary maps, define:

\begin{itemize}
    \item The group of \emph{\( p \)-cycles}:
    \[
    Z_p(K) = \ker(\partial_p),
    \]
    which consists of \( p \)-chains with zero boundary (closed cycles).

    \item The group of \emph{\( p \)-boundaries}:
    \[
    B_p(K) = \mathrm{im}(\partial_{p+1}),
    \]
    which consists of \( p \)-chains that are themselves boundaries of \((p+1)\)-chains.
\end{itemize}

The \emph{\( p \)-th simplicial homology group} is then defined as the quotient
\[
H_p(K) = Z_p(K) / B_p(K),
\]
which measures the \( p \)-cycles modulo those that bound higher-dimensional simplices. Intuitively, elements of \( H_p(K) \) correspond to \( p \)-dimensional “holes” in the space. Since $\mathbb{Z}_2$ is a field, each homology group forms a vector space.\\

\subsection{Zigzag Persistence}

Simplicial homology provides a powerful tool for analyzing the topology of a single, static simplicial complex. However, real-world data is often not a single static object but a sequence of related data sampled at different moments or corresponding to different parameters of evolution. In order to study these evolving objects, we define a 
\emph{zigzag filtration}. A zigzag filtration $\cZ$ is a sequence of simplicial complexes
\begin{equation*}
    \cZ : K_0 \subseteq K_1 \supseteq K_2 \subseteq \hdots \supseteq K_n.
\end{equation*}
where the sequence of inclusions could be both in forward and backward directions.  

The homology group of $K_i$ is a vector space $H_p(K_i)$ (under a field coefficient like $\mathbb{Z}_2$) for each $i\in\{1,\dots,n\}$. The inclusion map between $K_i$ and $K_j$, induces a natural linear map between $H_p(K_i)$ and $H_p(K_j)$. By assembling these vector spaces and linear maps together, we obtain the \emph{zigzag persistence module} $ M_{\cZ}$ defined by 
\begin{equation*}
 M_{\cZ}\!:\! H_p(K_0) \!\rightarrow \!H_p(K_1)\! \leftarrow  \dots \leftarrow H_p(K_n)
\end{equation*}

The Interval Decomposition Theorem for zigzag persistence modules \cite{carlsson2010zigzag, Gabriel1972}, states that $M_{\mathcal{Z}}$ decomposes uniquely (up to reordering and isomorphism) into interval modules:

\[ M_{\mathcal{Z}} \cong \bigoplus_{k} I[b_k, d_k] \]

where each $I[b_k, d_k]$ is an interval module with the support on the interval $[b_k,d_k]$, called a bar. 
% Formally an interval module \[I[b_k, d_k]: I_0 \!\rightarrow \!I_1\! \leftarrow I_2  \dots \leftarrow I_n\] is defined such as $I_i$ being the zero vector space for all $i$ except for $i\in [b_k, d_k]$,  for which $I_i=\mathbb{Z}_2$, and all linear maps are defined to be zero, except between two copies of $\mathbb{Z}_2$, for which the identity map is defined. 
The collection of these bars yields the \emph{zigzag barcode}, defined as the multiset of pairs $\{[b_k, d_k]\}_k$. This barcode carries the topological information present in the zigzag persistence module, which we leverage for hallucination detection.

One of the important properties of zigzag persistence is its stability under small perturbations \cite{carlsson2010zigzag}, implying that minor changes in attention weights between layers result in only small changes in the barcodes. This stability property ensures that the topological signatures we extract are robust to noise.

\subsection{Time Complexity}

We use the algorithm proposed in~\cite{fzz} to compute zigzag persistence barcodes. The authors show that the time complexity for computing zigzag persistent homology is the same as the time complexity for computing standard persistent homology. The time complexity for computing standard persistent homology is $O(n^\omega)$, where $n$ is the number of simplices~\cite{computing_pers_hom} and $\omega<2.371339$ is the matrix multiplication exponent \cite{doi:10.1137/1.9781611978322.63}. However, for the special case of graphs there exists a near linear time algorithm \cite{dey_hou_near_linear_zigzag} that could potentially be used. 

% To compute the zigzag barcode of the attention graphs, we used the python package $pyfzz$ from \cite{fzz}. The algorithm to compute persistent homology has a time complexity of $O(n^\omega)$, where $n$ is the number of simplices and $\omega<2.371339$ is the matrix multiplication exponent \cite{doi:10.1137/1.9781611978322.63}. However, for the special case of graphs there exists a near linear time algorithm \cite{dey2021computing} that could potentially be implemented. 

% \onecolumn

\section{Additional Experimental Details}
\label{app:exps}

To ensure the robustness of our method, we conducted several experiments to select the optimal hyperparameters for our main results. We use an NVIDIA A100 (40GB) GPU with 16 cores for all the experiments. 

\subsection{Hyperparameter Tuning}

\paragraph{Minimum Persistence Filtering.} As in well-known in the applied TDA community, shorter bars are considered to be noise. Hence, we filter out shorter bars and retain the longer ones. In order to choose the optimal threshold, we evaluate the performance after filtering out bars below various persistence thresholds (5, 7, 9 or 11 layers) at two different levels of graph sparsity (edge selection) - 30\% (Table~\ref{tab:layers1}) and 10\% (Table~\ref{tab:layers2}). Based on our results, we find that a moderate level of persistence filtering is optimal. Therefore, to balance the benefit of removing noise against the risk of discarding valuable signals, we select a conservative minimum persistence threshold of 5 for all main experiments. 

 % As we hypothesized that longer bars correspond to more cohesive arguments and truthful responses, we filtered out short bars and retained only those with high persistence. For Vicuna-7b, which has 32 layers, we defined high persistence as bars being present in 5, 7, 9, 11, or more layers, see Tables \ref{tab:layers1}, \ref{tab:layers2}. Bars with lower persistence were excluded prior to the vectorization of the persistence diagrams, ensuring that only the most structurally significant topological features were used.

\paragraph{Edge Selection Threshold.} The construction of our attention graphs depends on a sparsity parameter. We investigated the impact of this by varying the percentage of top attention weights retained (5\%, 10\%, 20\%, and 30\%) for different minimum persistence values - 5 (Table~\ref{tab:top_edges1}) and 9 (Table~\ref{tab:top_edges2}). The results indicate that retaining the top 10\% of edges provides a strong balance between detection performance and computational efficiency. We therefore use this 10\% threshold for all main experiments reported in the paper.

 % In addition, we investigated the impact of edge selection thresholds on the construction of attention graphs. Specifically, we varied the percentage of the most significant edges —those with the highest attention weights— by retaining the top 5\%, 10\%, 20\% and 30\% of the edges, the results are shown in Tables \ref{tab:top_edges1}, \ref{tab:top_edges2}. For further experimental results we retain only the top 10\% of edges, higher percentage of retention of edges may improve several metrics at the cost of higher computation time.
 %, we computed the correlation between the threshold and the top \% of edges in Table \ref{tab:pearson_coef}. \\

\subsection{Implementation Details}

\paragraph{Model Temperature.} We used a temperature value of 0.7 for all the models while generating answers. For GPT-4o-mini, we used a temperature of 0 while generating the annotation labels on QA Benchmarks.

\paragraph{Topological Feature Vectorization.} We use GUDHI~\cite{gudhi:PersistenceRepresentationsScikitlearnInterface} for all vectorization schemes. The specific parameters are as follows:

\begin{itemize}
    \item \textbf{Persistence Images (\texttt{PersImg)}:} We use a resolution of \texttt{32 x 32}.
    \item \textbf{Betti Curves:} We use a sampling resolution of \texttt{32} points.
    \item \textbf{Persistence Entropy (\texttt{PersEntropy}):} We use the default implementation from the library.
\end{itemize}

\paragraph{Classifier.} For all our experiments, we classify using a random forest classifier. For experiments on RAGTruth Dataset, we use \texttt{n\_estimators = 500}. For all other experiments, we use \texttt{n\_estimators = 100}. The \texttt{max\_depth} parameter was tuned for each configuration. The values are reported in Table~\ref{app:tab:rf_hyperparam}.

\paragraph{Baseline Reproduction.} To ensure a fair comparison with LapEigVals baseline from~\cite{binkowski2025hallucinationdetectionllmsusing}, we followed their experimental protocol precisely. As specified in their work, we retained the top $k=50$ eigenvalues across all heads and layers and trained a logistic regression classifier with their reported parameters.

 % For  all the datasets, we classify using a random forest classifier with \texttt{n\_estimators = 100}. We report the values of \texttt{max\_depth} we used for different configurations in Table~\ref{app:tab:rf_hyperparam}. We use a resolution of $32 \times 32$ for generating Persistence Images (\texttt{PersImg}). We use a resolution of $32$ for generating Betti Curves (\texttt{Betti Curve}) for all datasets. We refer the reader to the documentation of these methods in~\cite{gudhi:PersistenceRepresentationsScikitlearnInterface} for further details about these methods. For generating the value of Persistence Entropy, we did not use any additional hyperparameter. We used the implementation in~\cite{gudhi:PersistenceRepresentationsScikitlearnInterface} as is. 

% To reproduce the results of ~\cite{binkowski2025hallucinationdetectionllmsusing} in Table \ref{tab:qa_based_detection}, we retained the top $k=50$ eigenvalues, and used all heads and layers, we then trained a logistic regression classifier with the same parameters as mentioned in \cite{binkowski2025hallucinationdetectionllmsusing}.

% We used FP32 precision for all datasets, except for RAGTruth, where FP16 was used to improve memory efficiency.

\subsection{Detailed Performance Metrics and Statistical Uncertainty}

To ensure reproducibility of our results, this section presents the detailed performance metrics for \topoframework{} across all datasets. The main results reported in the body of the paper correspond to the single best run for clarity. Here, we report the mean and standard deviation calculated over five independent runs with different random seeds. Refer to Table~\ref{app:tab:fava_mean_std} and Table~\ref{app:tab:qa_mean_std}. For the RAGTruth dataset, we use the official train-test split provided in the dataset. Consequently, all results reported in the main text are based on this single, pre-defined partition. For all other datasets, we use a 80-20 train-test split.

\begin{table}[!htbp]
  \centering
  \setlength{\tabcolsep}{6pt}
  \resizebox{\columnwidth}{!}{
  \begin{tabular}{l|l|ccc}
    \toprule
    \textbf{Dataset} & \textbf{Model} & \textbf{\texttt{PersImg}} & \textbf{\texttt{PersEntropy}} & \textbf{\texttt{BettiCurve}} \\
    \midrule

    % ------------------ FAVA Annotated -------------------
    \multirow{3}{*}{\textbf{FAVA Annotated}} 
      & \texttt{Llama-2-7b}   & 20  & 10  &  10 \\
      & \texttt{Llama-3-8b}   & 20  &  10 & 10  \\
      & \texttt{Vicuna-7b}    & 20  & 10  &  10 \\
    \midrule

    % ------------------ RAGTruth -------------------
    \multirow{3}{*}{\textbf{RAGTruth}} 
      & \texttt{Llama-2-7b}   & 25  & 25  & 25  \\
      & \texttt{Llama-2-13b}  &  25 &  25 &  25 \\
      & \texttt{Mistral-7b}   &  25 & 25  & 25  \\
    \midrule

    % ------------------ NQOpen -------------------
    \multirow{2}{*}{\textbf{NQOpen}} 
      & \texttt{Llama-3.1-8b} & 6 & 4  & 5  \\
      & \texttt{Llama-3.2-3b} & 5  & 4 &  4 \\
    \midrule

    % ------------------ TruthfulQA -------------------
    \multirow{2}{*}{\textbf{TruthfulQA}} 
      & \texttt{Llama-3.1-8b} & 6  & 5  & 5  \\
      & \texttt{Llama-3.2-3b} & 6  & 4  &  5 \\
    \bottomrule
  \end{tabular}
  }
  \caption{Maximum tree depth (\texttt{max\_depth}) used in Random Forest Classifier.}
  \label{app:tab:rf_hyperparam}
\end{table}

\begin{table}[h!]

\centering
\resizebox{\columnwidth}{!}{
  \begin{tabular}{|c|c|c|c|c|c|}
    \toprule
    \multirow{1}{*}{\textbf{Method}} & \multirow{1}{*}{\textbf{Metric}} & \multicolumn{4}{c|}{\textbf{Min. Persistence of Bars}}  \\
     \cmidrule(lr){3-6}
    & & $\geqslant 5$ & $\geqslant 7$  & $\geqslant 9$ & $\geqslant 11$ \\
    
    \midrule
    \multirow{4}{*}{\texttt{PersImg}}
    &AUC-ROC        & 73.46 & 73.27          & 72.93 & \textbf{73.76} \\
    &Accuracy       & 65.22 & \textbf{68.48} & 65.22 & 64.13 \\
    &TPR at 5\% FPR & 7.41  & 7.41           & 7.41  & 7.41  \\
    &F1 Score       & 72.88 & \textbf{75.63} &\textbf{ 72.88 }& 72.27\\
    \midrule
    \multirow{4}{*}{\texttt{PersEntropy}}
 & AUC-ROC        & 75.41 & \textbf{77.75} & \textbf{74.44} & 69.98 \\
 & Accuracy       & 66.30 & 68.48 & 65.22 & \textbf{65.22 }\\
 & TPR at 5\% FPR & \textbf{33.33} & 25.93 & 24.07 & 22.22 \\
 & F1 Score       & 71.56 & 74.34 & 70.91 & 72.41\\
    \midrule
    \multirow{4}{*}{\texttt{Betti Curve}}
 & AUC-ROC        & \textbf{75.78} & 73.73 & 72.59 & \textbf{74.29} \\
 & Accuracy       & \textbf{69.57} & 67.39 & 64.13 & \textbf{65.22} \\
 & TPR at 5\% FPR & 27.78 & \textbf{29.63} & \textbf{29.63} & \textbf{37.04} \\
 & F1 Score       & \textbf{75.86} & 74.14 & 71.30 & \textbf{73.77}\\
    \bottomrule
  \end{tabular}
  }
\caption{This table evaluates our three vectorization schemes when filtering out topological features (bars) with low persistence. We select the top 30\% of the edges for this experiment.}
\label{tab:layers1}
\end{table}

% \onecolum

\begin{table}[h!]

\centering
\resizebox{\columnwidth}{!}{
  \begin{tabular}{|c|c|c|c|c|c|}
    \toprule
    \multirow{1}{*}{\textbf{Method}} & \multirow{1}{*}{\textbf{Metric}} & \multicolumn{4}{c|}{\textbf{Min. Persistence of Bars}}  \\
     \cmidrule(lr){3-6}
    & & $\geqslant 5$ & $\geqslant 7$  & $\geqslant 9$ & $\geqslant 11$ \\
    
    \midrule
    \multirow{4}{*}{\texttt{PersImg}}
 & AUC-ROC        & 73.42 & 73.08 & 72.34 & 74.10 \\
 & Accuracy       & 66.30 & 64.13 & 64.13 & 68.48 \\
 & TPR at 5\% FPR & 12.96 & 25.93 & 14.81 & 20.37 \\
 & F1 Score       & 73.95 & 71.79 & 71.79 & 76.03\\
    \midrule
    \multirow{4}{*}{\texttt{PersEntropy}}
 & AUC-ROC        & 66.96 & 57.36 & 59.23 & 64.91 \\
 & Accuracy       & 65.22 & 64.13 & 60.87 & 65.22 \\
 & TPR at 5\% FPR & 7.41  & 7.41  & 14.81 & 14.81 \\
 & F1 Score       & 70.91 & 70.80 & 67.27 & 71.93\\
    \midrule
    \multirow{4}{*}{\texttt{Betti Curve}}
 & AUC-ROC        & 69.96 & 70.49 & 71.03 & 72.69 \\
 & Accuracy       & 66.30 & 65.22 & 64.13 & 65.22 \\
 & TPR at 5\% FPR & 12.96 & 14.81 & 20.37 & 18.52 \\
 & F1 Score       & 73.50 & 72.41 & 71.79 & 72.88\\
    \bottomrule
  \end{tabular}
  }
\caption{This table evaluates our three vectorization schemes when filtering out topological features (bars) with low persistence. We select the top 10\% of the edges for this experiment.}
\label{tab:layers2}
\end{table}

\begin{table}[h!]

\centering
\resizebox{\columnwidth}{!}{
  \begin{tabular}{|c|c|c|c|c|c|}
    \toprule
    \multirow{1}{*}{\textbf{Method}} & \multirow{1}{*}{\textbf{Metric}} & \multicolumn{4}{c|}{\textbf{Top \% of edges (attn weights) selected}}  \\
     \cmidrule(lr){3-6}
    & & 30\% & 20\% & 10\%& 5\% \\
    
    \midrule
    \multirow{4}{*}{\texttt{PersImg}}
 & AUC-ROC        & 73.46 & 72.83 & 73.42 & 71.47 \\
 & Accuracy       & 65.22 & 66.30 & 66.30 & 65.22 \\
 & TPR at 5\% FPR & 7.41  & 20.37 & 12.96 & 9.26  \\
 & F1 Score       & 72.88 & 73.50 & 73.95 & 72.88\\
    \midrule
    \multirow{4}{*}{\texttt{PersEntropy}}
 & AUC-ROC        & 75.41 & 74.39 & 66.96 & 74.78 \\
 & Accuracy       & 66.30 & 66.30 & 65.22 & 67.39 \\
 & TPR at 5\% FPR & 33.33 & 25.93 & 7.41  & 22.22 \\
 & F1 Score       & 71.56 & 71.56 & 70.91 & 74.14 \\
    \midrule
    \multirow{4}{*}{\texttt{Betti Curve}}
 & AUC-ROC        & 75.78 & 72.78 & 69.96 & 74.17 \\
 & Accuracy       & 69.57 & 66.30 & 66.30 & 65.22 \\
 & TPR at 5\% FPR & 27.78 & 31.48 & 12.96 & 38.89 \\
 & F1 Score       & 75.86 & 73.95 & 73.50 & 71.93\\
    \bottomrule
  \end{tabular}
  }
\caption{This table shows the performance of our three vectorization schemes while varying the percentage of top attention weights used to construct the attention graphs (minimum bar persistence is held constant at 5). }
\label{tab:top_edges1}
\end{table}

\begin{table}[h!]

\centering
\resizebox{\columnwidth}{!}{
  \begin{tabular}{|c|c|c|c|c|c|}
    \toprule
    \multirow{1}{*}{\textbf{Method}} & \multirow{1}{*}{\textbf{Metric}} & \multicolumn{4}{c|}{\textbf{Top \% of edges (attn weights) selected }}  \\
     \cmidrule(lr){3-6}
    & & 30\% & 20\% & 10\% & 5\% \\
    
    \midrule
    \multirow{4}{*}{\texttt{PersImg}}
 & AUC-ROC        & 72.34 & 72.93 & 72.39 & 72.00 \\
 & Accuracy       & 64.13 & 65.22 & 66.30 & 65.22 \\
 & TPR at 5\% FPR & 14.81 & 7.41  & 16.67 & 18.52 \\
 & F1 Score       & 71.79 & 72.88 & 73.95 & 72.88 \\
    \midrule
    \multirow{4}{*}{\texttt{PersEntropy}}
 & AUC-ROC        & 74.44 & 71.08 & 59.23 & 74.98 \\
 & Accuracy       & 65.22 & 64.13 & 60.87 & 67.39 \\
 & TPR at 5\% FPR & 24.07 & 14.81 & 14.81 & 25.93 \\
 & F1 Score       & 70.91 & 71.30 & 67.27 & 73.21\\
    \midrule
    \multirow{4}{*}{\texttt{Betti Curve}}
 & AUC-ROC        & 72.59 & 72.39 & 71.03 & 72.64 \\
 & Accuracy       & 64.13 & 66.30 & 64.13 & 66.30 \\
 & TPR at 5\% FPR & 29.63 & 22.22 & 20.37 & 33.33 \\
 & F1 Score       & 71.30 & 73.95 & 71.79 & 72.07\\
    \bottomrule
  \end{tabular}
  }
\caption{This table shows the performance of our three vectorization schemes while varying the percentage of top attention weights used to construct the attention graphs (minimum bar persistence is held constant at 9). }
\label{tab:top_edges2}
\end{table}

\begin{table}[!htbp]
  \centering
  \setlength{\tabcolsep}{6pt}
  \resizebox{\columnwidth}{!}{
  \begin{tabular}{l|cc|cc}
    \toprule
    \multirow{2}{*}{\textbf{Method}} &
    \multicolumn{2}{c|}{\textbf{TruthfulQA (AUC-ROC)}} &
    \multicolumn{2}{c}{\textbf{NQOpen (AUC-ROC)}} \\
    \cmidrule(lr){2-5}
    & \texttt{Llama-3.1-8b} & \texttt{Llama-3.2-3b} &
      \texttt{Llama-3.1-8b} & \texttt{Llama-3.2-3b} \\
    \midrule
    \texttt{PersImg}      & $65.92 \pm 4.05$ & $62.08 \pm 2.32$ & $68.29 \pm 2.58$ & $67.27 \pm 2.16$ \\
    \texttt{PersEntropy}  & $63.62 \pm 3.07$ & $58.34 \pm 6.61$ & $66.11 \pm 1.36$ & $65.53 \pm 1.68$ \\
    \texttt{BettiCurve}   & $59.31 \pm 6.11$ & $58.22 \pm 4.74$ & $67.97 \pm 2.92$ & $65.86 \pm 2.07$ \\
    \bottomrule
  \end{tabular}
  }
  \caption{Each entry reports mean $\pm$ standard deviation across five random seeds for Test AUC-ROC scores.}
  \label{app:tab:qa_mean_std}
\end{table}

\begin{table*}[!htbp]
  \centering
  \renewcommand{\arraystretch}{1.15}
  \setlength{\tabcolsep}{5pt}
  \resizebox{\textwidth}{!}{
  \begin{tabular}{l|ccc|ccc|ccc}
    \toprule
    \multirow{2}{*}{\textbf{Metric}} & 
    \multicolumn{3}{c|}{\textbf{\texttt{PersImg}}} & 
    \multicolumn{3}{c|}{\textbf{\texttt{PersEntropy}}} & 
    \multicolumn{3}{c}{\textbf{\texttt{Betti Curve}}} \\
    \cmidrule(lr){2-10}
    & \texttt{Llama-2-7b} & \texttt{Llama-3-8b} & \texttt{Vicuna-7b} &
      \texttt{Llama-2-7b} & \texttt{Llama-3-8b} & \texttt{Vicuna-7b} &
      \texttt{Llama-2-7b} & \texttt{Llama-3-8b} & \texttt{Vicuna-7b} \\
    \midrule
    \textbf{AUC-ROC} &
      $78.72 \pm 3.41$ & $78.89 \pm 2.74$ & $79.20 \pm 3.38$ &
      $70.35 \pm 4.53$ & $72.29 \pm 4.34$ & $71.31 \pm 3.63$ &
      $71.64 \pm 3.75$ & $72.71 \pm 4.90$ & $72.95 \pm 2.48$ \\
    \textbf{Accuracy} &
      $73.04 \pm 2.70$ & $74.35 \pm 2.24$ & $74.13 \pm 2.70$ &
      $67.83 \pm 2.54$ & $71.52 \pm 2.52$ & $69.57 \pm 3.30$ &
      $67.83 \pm 3.12$ & $71.30 \pm 2.54$ & $70.65 \pm 2.75$ \\
    \textbf{TPR @ 5\% FPR} &
      $30.02 \pm 12.17$ & $37.70 \pm 11.57$ & $36.62 \pm 12.25$ &
      $19.86 \pm 3.70$ & $16.97 \pm 4.61$ & $25.55 \pm 8.08$ &
      $21.41 \pm 3.95$ & $16.45 \pm 6.69$ & $22.76 \pm 8.28$ \\
    \textbf{F1 Score} &
      $80.21 \pm 1.66$ & $81.42 \pm 1.77$ & $81.16 \pm 1.70$ &
      $76.77 \pm 2.16$ & $78.15 \pm 2.33$ & $76.72 \pm 2.53$ &
      $77.05 \pm 2.54$ & $79.45 \pm 1.60$ & $78.96 \pm 2.30$ \\
    \bottomrule
  \end{tabular}
  }
  \caption{ 
  Each entry reports mean $\pm$ standard deviation across five random seeds for all the metrics on the FAVA Annotated Dataset.}
  \label{app:tab:fava_mean_std}
\end{table*}

\newpage
\onecolumn
\section{QA prompt}
We used OpenAI’s API and ChatGPT-4o-mini model as judge for labeling the generated response of each LLM in the whitebox setting. The prompt we used is the same for NQopen, and TruthfullQA.

\begin{tcolorbox}[
    colback= white,
    colframe=black,
    boxrule=0.5pt,
    sharp corners,
    title= Prompt used for LLM-as-judge with Open AI's ChatGPT4o-mini,
    fonttitle=\bfseries,
    breakable]

\noindent You are an evaluator for QA tasks. 
Given the question, the reference best answer, and a model's answer, classify the model's answer as: \begin{itemize}
    \item[-] CORRECT if it matches or is equivalent to the best answer.
    \item[-] HALLUCINATION if it introduces facts not in the best answer.
\item[-] N/A if the model expresses uncertainty or says it cannot answer.\\
Respond with only one label: CORRECT, HALLUCINATION, or N/A.

Question: \{question\}\\
Reference best answer: \{best\_answer\}\\
Model answer: \{model\_answer\}\\

\end{itemize}
\end{tcolorbox}

\end{document}